\PassOptionsToPackage{numbers,sort&compress}{natbib}

\documentclass{article}

\usepackage[preprint]{neurips_2025}
\usepackage[utf8]{inputenc}
\usepackage[T1]{fontenc}
\usepackage{hyperref}
\usepackage{url}
\usepackage{booktabs}
\usepackage{amsfonts}
\usepackage{nicefrac}
\usepackage{microtype}
\usepackage{xcolor}

\usepackage[normalem]{ulem}

\usepackage{amsmath}
\usepackage{graphicx}
\usepackage{algorithm}
\usepackage{algpseudocode}

\setcitestyle{square,comma}

\title{Combined Representation and Generation with Diffusive State Predictive Information Bottleneck}

\author{
  Richard John \\
  Department of Physics\thanks{College Park, MD, 20742} \\
  Institute for Physical Science and Technology\footnotemark[1] \\
  University of Maryland \\
  \And
  Yunrui Qiu \\
  Institute for Physical Science and Technology\footnotemark[1] \\
  Institute for Health Computing\thanks{Bethesda, MD, 20852} \\
  University of Maryland \\
  \AND
  Lukas Herron \\
  Biophysics Program\footnotemark[1] \\
  Institute for Physical Science and Technology\footnotemark[1] \\
  Institute for Health Computing\footnotemark[2] \\
  University of Maryland \\
  \And
  Pratyush Tiwary\thanks{Email: ptiwary@umd.edu} \\
  Department of Chemistry and Biochemistry\footnotemark[1] \\
  Institute for Physical Science and Technology\footnotemark[1] \\
  Institute for Health Computing\footnotemark[2] \\
  University of Maryland \\
  \texttt{} \\
}

\begin{document}

\maketitle

\begin{abstract}
Generative modeling becomes increasingly data-intensive in high-dimensional spaces. In molecular science, where data collection is expensive and important events are rare, compression to lower-dimensional manifolds is especially important for various downstream tasks, including generation. We combine a time-lagged information bottleneck designed to characterize molecular important representations and a diffusion model in one joint training objective. The resulting protocol, which we term Diffusive State Predictive Information Bottleneck (D-SPIB), enables the balancing of representation learning and generation aims in one flexible architecture. Additionally, the model is capable of combining temperature information from different molecular simulation trajectories to learn a coherent and useful internal representation of thermodynamics. We benchmark D-SPIB on multiple molecular tasks and showcase its potential for exploring physical conditions outside the training set.
\end{abstract}

\section{Motivation} \label{sec:motivation}

In various scientific areas, probability distributions of interest are inaccessible and characterized only by a set of known samples. Biomolecules like proteins, for example, are highly dynamic and can adopt diverse conformations that critically influence their biological functions. However, sampling the equilibrium distribution of protein structure is usually prohibitively costly, and even when simulation is feasible, it only yields the structure ensemble at a particular thermodynamic state (e.g., a certain temperature) \citep{lindorff2011fast, noe2019boltzmann, dibak2022temperature}. Generating samples from the equilibrium distribution is especially valuable when training data is scarce or difficult to obtain \citep{tiwary2024generative, lewis2025scalable}. This goal is addressed by a broad class of machine learning algorithms known as probabilistic generative models (PGMs). Motivated by widespread, successful application of PGMs to sampling problems, and the Information Bottleneck (IB) principle to learning physically-motivated representations, we introduce the Diffusive State Predictive Information Bottleneck (D-SPIB) and examine its performance on molecular modeling tasks. In addition, we describe how D-SPIB may infer the temperature dependence of metastable states from limited multi-temperature data. Temperature-aware generative models have been introduced in contemporary works, where they have been demonstrated to enhance sampling post-simulation \cite{herron2024inferring,lee2025exponentially,beyerle2025inferring,qiu2025latent}.

\textbf{Generative modeling in lower-dimensional spaces is advantageous.} In machine learning generally, the `curse of dimensionality' refers to a phenomenon where the amount of data required to train a model scales exponentially with data dimensionality. When the amount of training data is insufficient, the model overfits and exhibits undesirable behavior like memorization \citep{tsybakov2009nonparametric}. Indeed, theoretical study has identified a critical `collapse time' in diffusion models--a popular class of PGMs--which indicates memorization of the training set \citep{biroli2024dynamical}. The phenomenon is corroborated by practical experiments, which have demonstrated that generative models become less accurate at density estimation as dimensionality increases when the amount of training data is held constant \citep{john2025comparison}.

\textbf{State Predictive Information Bottleneck (SPIB) provides a physically meaningful reduced-dimensional space.} This physics-inspired framework of the variational IB finds a set of representations, i.e., collective variables, 
for molecular systems that capture the important slowest degrees of freedom and provide maximal information for dynamics propagation \citep{wang2021state, wang2024information}. The representations derived from SPIB have been validated on tasks such as molecular kinetics calculation \citep{lee2024calculating} and exploration of aqueous crystal nucleation \citep{wang2024local}. Recently, the Latent Thermodynamic Flows (LaTF) model was developed to simultaneously train a normalizing flow PGM and an SPIB model for unified generative modeling and representation learning \citep{qiu2025latent}.

\textbf{Jointly learning to represent and generate data outperforms generation with pre-trained encodings.} Variational autoencoders (VAEs) \citep{kingma2013auto} aim to find a low-dimensional latent embedding from which the data may be reconstructed. As noted above, these latent spaces are strong candidates for generative modeling \citep{rombach2022high, preechakul2022diffusion}. Contemporary work indicates that simultaneous optimization of a generative model and latent variables outperforms serial optimization on reconstructive tasks \citep{liu2024unified, qiu2025latent}.

Based on the above criteria, we believe diffusion models and the SPIB latent space as implemented in the proposed D-SPIB framework constitute a suitable pairing for joint representation learning and generative modeling.

\section{D-SPIB} \label{sec:architecture}
\begin{figure}[ht]
    \centering
    \includegraphics[width=0.62\textwidth]{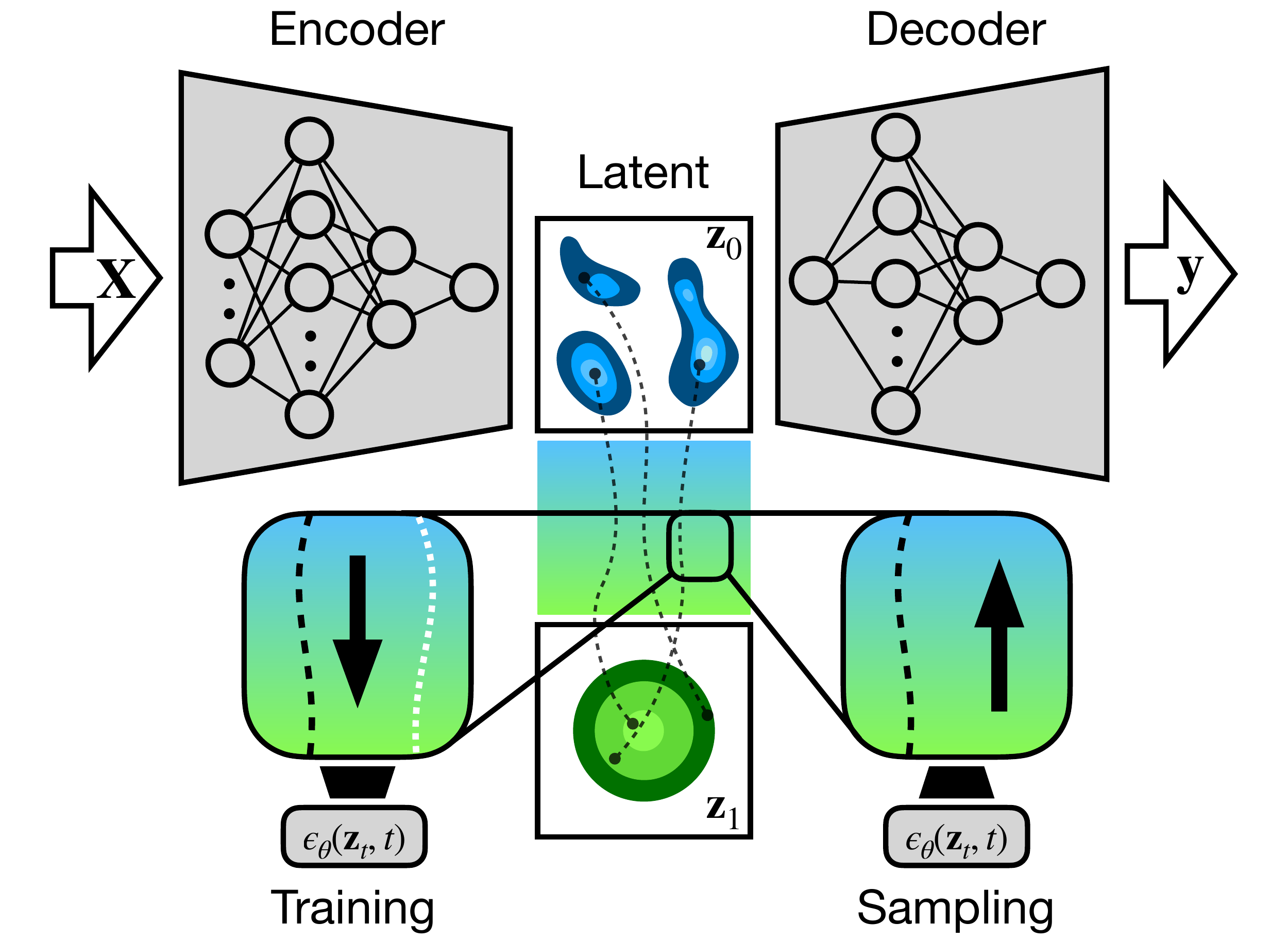}
  \caption{Diffusive SPIB architecture. Input data $\mathbf{X}$ is encoded to latent $\mathbf{z}_0$ and decoded to state label $\mathbf{y}$. The distribution of the encoded variable $\mathbf{z}_0$ is regularized by the IB prior distribution generated from $\mathbf{z}_1$ using a diffusion model with an easily sampled, pre-defined generative prior distribution.
  Blow-ups show the diffusion trajectories, including the reference forward trajectories used for training (white), the learned forward trajectories (black), and the backward trajectories employed for sampling (black).}
  \label{fig: NN Diagram}
\end{figure}

The SPIB architecture consists of a time-lagged VAE trained to predict the state of a molecular system at future time $t+\tau$ based on high-dimensional molecular embeddings at time $t$. Increasingly expressive priors for the VAE are sought after as development of SPIB continues \citep{tomczak2018vae, wang2021state}. Accordingly, we develop the D-SPIB model by incorporating a score-based generative model to generate a more flexible and expressive IB prior distribution that closely aligns with the SPIB-encoded distribution (see the architecture in Fig. \ref{fig: NN Diagram}).

The SPIB loss function is a variational approximation to the IB principle \citep{tishby2000information}:
\begin{align}
\label{eq:spib}
    \mathcal{L}_{\mathrm{SPIB}}
    = \frac{1}{N}\sum_{n=1}^{N}\int d \mathbf{z} \, p_{\theta}(\mathbf{z}|\mathbf{X}^{n})\Big[-\log q_{\theta}(\mathbf{y}^{n+\tau}|\mathbf{z}) + \beta \log\frac{p_{\theta}(\mathbf{z}|\mathbf{X}^{n})}{r_{\theta}(\mathbf{z})}\Big]
\end{align}

where $p_{\theta}(\mathbf{z}|\mathbf{X}^{n})$ is an encoder that maps molecular embeddings $\mathbf{X}^n \in \mathbb{R}^D$ to latent variable $\mathbf{z} \in \mathbb{R}^d$ and $q_{\theta}(\mathbf{y}^{n+\tau}|\mathbf{z})$ is a decoder that predicts future-transition-state label $\mathbf{y}^{n+\tau} \in \mathbb{R}$ from $\mathbf{z}$. The lag time $\tau$ enables the model to learn representations relevant for long-term dynamics, while the regularization factor $\beta$ constrains the encoded distribution $p_{\theta}(\mathbf{z}|\mathbf{X}^{n})$ to remain close to the prior $r_{\theta}(\mathbf{z})$.

We present the D-SPIB objective function (derived in full in Appendix \ref{sec:derivation}):
\begin{align}
\label{eq:dspib}
    &\mathcal{L}_{\mathrm{D-SPIB}} = \frac{1}{N}\sum_{n=1}^{N} \Big[\mathbb{E}_{p_{\theta}(\mathbf{z}_0|\mathbf{X}^{n})} \Big[-\log q_{\theta}(\mathbf{y}^{n+\tau}|\mathbf{z}_0) + \beta \log p_{\theta}(\mathbf{z}_0|\mathbf{X}^{n})\Big] \notag \\
    & + \beta \cdot \mathbb{E}_{t \sim \mathcal{U}[0,1]} \Big[ \frac{g(t)^2}{2} \mathbb{E}_{p_{\theta}(\mathbf{z}_t, \mathbf{z}_0 | \mathbf{X}^n)} \Big[ || \nabla_{\mathbf{z}_t} \log p(\mathbf{z}_t | \mathbf{z}_0) - \nabla_{\mathbf{z}_t} \log r_{\theta}(\mathbf{z}_t) ||^2_2 \Big] \Big] \Big]
\end{align}

In place of SPIB’s standard prior, we employ a diffusion model to construct a more flexible and trainable prior distribution. Following Theorem 1 of \citep{vahdat2021score}, the regularization term in Eq. \ref{eq:spib} can be written as a cross-entropy and further reformulated as a score-matching objective. Here, the latent variable $\mathbf{z}_t \equiv \mathbf{z}(t)$ is a time-dependent variable ($t \in [0,1]$), where $\mathbf{z}_0$ denotes the SPIB-encoded variable and $\mathbf{z}_1$ the diffusion prior variable. The diffusion dynamics for $\mathbf{z}$ follow stochastic differential equations (SDEs):

\begin{align}
    \textrm{Forward: }d\mathbf{z} &= f(\mathbf{z},t) dt + g(t) d\mathbf{w} 
    \label{eq:forward} \\
    \textrm{Reverse: }d\mathbf{z} &= \big[ f(\mathbf{z},t) - g^2(t) \nabla_{\mathbf{z}} \log p_t(\mathbf{z}_t) \big] dt + g(t) d\mathbf{\bar{w}}
    \label{eq:reverse}
\end{align}

where $\mathbf{w}$ is the standard Wiener process. In the reverse equation, $dt$ and the Wiener process $\mathbf{\bar{w}}$ are understood to run from $t=1$ to $t=0$. Coupled SDEs like the above are at the foundation of score-based models \citep{song2020score}. In broad strokes, the forward process transports samples from an unknown density $p_0(\mathbf{z}_0)$ to a Gaussian prior distribution $p_1(\mathbf{z}_1)$. The stochastic transport induces a series of marginal densities $p_t(\mathbf{z}_t)$ that determine the score, $ \nabla_\mathbf{z} \log p_t(\mathbf{z}_t)$. Eq. \ref{eq:dspib} regresses $\nabla_{\mathbf{z}_t} \log r_{\theta}(\mathbf{z}_t)$ to the gradient of the marginal noising kernel $p(\mathbf{z}_t | \mathbf{z}_0)$, which is equivalent to the score and is entirely determined by the (pre-determined) functional forms of $f(t)$ and $g(t)$ \citep{vincent2011connection}. Once the score $\nabla_{\mathbf{z}_t} \log r_{\theta}(\mathbf{z}_t)$ is learned, the reverse diffusion process (Eq. \ref{eq:reverse}) can be simulated from randomly sampled $\mathbf{z}^*\sim p_1(\mathbf{z}_1)$ to generate samples in the latent space. Incorporating the diffusion model does not alter the IB principle, since $r_{\theta}(\mathbf{z})$ in Eq. \ref{eq:spib} is simply modeled as $r_{\theta}(\mathbf{z}_0)$ in Eq. \ref{eq:dspib}.

We further extend the representational capacity of D-SPIB by tempering the generative model, i.e., by adjusting the variance of the generative prior as a linear function of temperature \citep{noe2019boltzmann, dibak2022temperature, herron2024inferring, wang2022data}. During training, the variance of the prior is scaled on a per-sample basis based on an associated temperature value. We also construct a temperature embedding and inject it into the denoising network via residual-like additions, thereby making the score explicitly temperature dependent (see more details in Appendix \ref{Algorithms} and \ref{Experimental Details}). Tempering the generative prior in this way allows the D-SPIB model to learn the temperature dependence of the SPIB state populations and optimize the latent representation accordingly.

\section{Results} \label{sec:results}

\subsection{Benchmarking D-SPIB on an Analytical Potential System} \label{sec:single temp}

\begin{figure}[ht]
    \centering
    \includegraphics[width=0.8\textwidth]{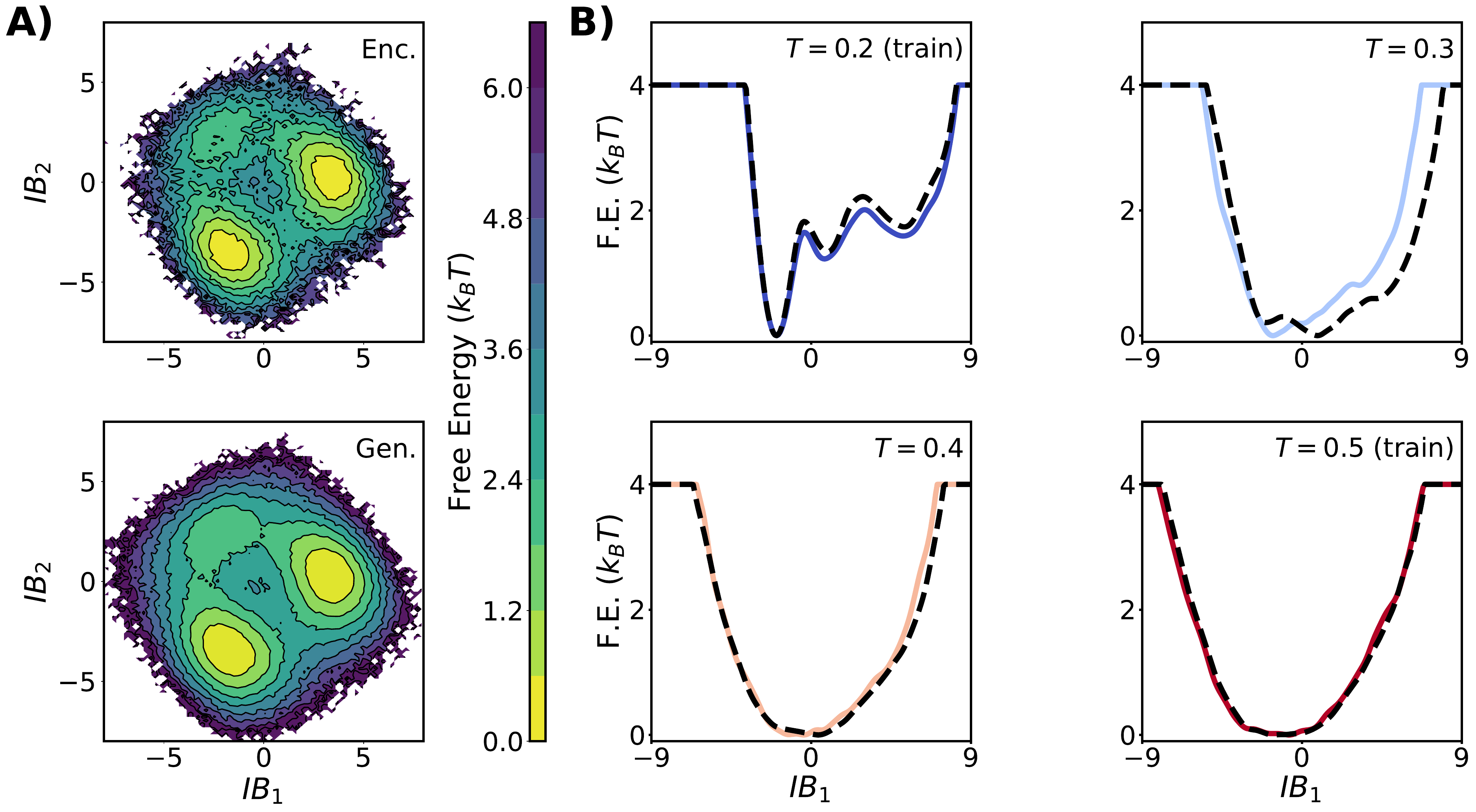}
  \caption{A) The distribution of encoded validation data and generated distribution by D-SPIB is shown for the three-hole potential. B) Free energy profiles along a D-SPIB latent dimension for generated data (colored, solid) and molecular dynamics data (black, dashed) in the multi-temperature LJ7 experiment.
  }
  \label{fig:FES Results}
\end{figure}

Our first experiment evaluates D-SPIB on a `three-hole' potential in two dimensions \citep{metzner2006illustration}. A trajectory of a single particle was generated by simulating Langevin dynamics for $5\times 10^{7}$ steps at temperature $T=1/k_B$, and a randomly-selected fifth of the simulation trajectory segment was removed to form a validation set. A SPIB model was trained on the simulation data, then the D-SPIB model was optimized (see Appendix \ref{Experimental Details}). We compare the D-SPIB generated free energy landscape in latent space to the D-SPIB encoded validation set landscape in Fig. \ref{fig:FES Results}. D-SPIB recovers the three expected metastable states with visual agreement between the distributions. Table \ref{table:kl-divergence} quantifies the agreement via the symmetrized KL divergence between the encoded and D-SPIB generated distributions (with standard deviation over three runs shown); the divergence of a generic SPIB model is provided for reference, where generated data is probabilistically drawn from the vanilla SPIB analytical prior \citep{wang2021state}, without any generative model.

\begin{table}
  \caption{Single Temperature $D_{KL}$}
  \label{table:kl-divergence}
  \centering
  \begin{tabular}{lll}
    \toprule
    System & $D_{KL}$ (SPIB) & $D_{KL}$ (D-SPIB) \\
    \midrule
    Three-hole & $9.773 \pm 0.011$ & $\mathbf{0.042 \pm 0.002}$ \\
    LJ7 ($T=0.2$) & $15.76 \pm 0.321$ & $\mathbf{0.760 \pm 0.123}$ \\
    \bottomrule
  \end{tabular}
\end{table}

\subsection{Application of D-SPIB to Lennard-Jones Particle System} \label{sec:multi temp}

Our second experiment concerns a two-dimensional seven-particle system with interactions governed by the Lennard–Jones potential (LJ7 system)\citep{tribello2010self, schwerdtfeger2024100}. Simulations were performed across temperatures $T \in \{0.2, 0.3, 0.4, 0.5\}$ (in units of $\epsilon / k_B$). We first show numerical results comparing D-SPIB and generic SPIB for the single temperature case ($T = 0.2$) in Table \ref{table:kl-divergence} (with standard deviation over three independent runs reported), where we observe the better generation performance of the D-SPIB architecture on the symmetrized KL divergence test.

To examine model accuracy in generating data under new, unseen thermodynamic conditions, we train at pre-selected temperatures $T \in \{ 0.2, 0.5\}$ and compare to intermediate temperatures where no data was provided. In Fig. \ref{fig:FES Results}B, we compare the distribution generated by D-SPIB to a reference distribution obtained via molecular dynamics simulations across four temperatures: two that the model was trained on and two intermediate values, finding good agreement in each case. D-SPIB correctly predicts that the minima of the LJ7 system thermalize suddenly between $T=0.2$ and $0.3$ and then more gradually from $T=0.3$ to $0.5$ (see Fig. \ref{fig:FES Results}B).

\section{Conclusion}

In this work we have presented D-SPIB as a generative model with state-predictive capacity inherited from SPIB. We re-expressed the prior term in the SPIB objective function to derive the D-SPIB objective, and explained how the temperature dependence of the free energy may be inferred by learning temperature embedding and tempering the generative prior. D-SPIB's generative capabilities were evaluated on an analytic three-hole potential, where D-SPIB achieved a lower divergence between the encoded and generated latent distributions compared to SPIB. We further examined how well D-SPIB predicts the temperature dependence of the Boltzmann distribution of the LJ7 cluster of particles, and found that D-SPIB predicts rapid thermalization between temperatures of 0.2 and 0.3, which is supported by molecular dynamics simulation. Overall, we believe D-SPIB to be a suitable model for learning both informative representations and modeling the temperature dependence of systems in a highly data-efficient manner.

\section{Code availability} \label{sec:code}

The code containing the model, datasets, experiments, and analysis is available at the following GitHub link: \url{https://github.com/rickyjohnwhu/diff-spib}.

\section{Acknowledgments}

This research was entirely supported by the US Department of Energy, Office of Science, Basic Energy Sciences, CPIMS Program, under Award No. DE-SC0021009. We thank UMD HPC’s Zaratan and NSF ACCESS (project CHE180027P) for computational resources. P.T. is an investigator at the University of Maryland Institute for Health Computing, which is supported by funding from Montgomery County, Maryland and The University of Maryland Strategic Partnership: MPowering the State, a formal collaboration between the University of Maryland, College Park, and the University of Maryland, Baltimore.

\clearpage

\bibliography{ref}

\clearpage

\appendix

\section{Loss function derivation}
\label{sec:derivation}

We derive the D-SPIB objective function beginning with the generic SPIB loss function and using Thm. 1 of \citep{vahdat2021score} to express the prior in terms of a diffusion loss.

The SPIB loss function is as follows:

\begin{align}
    \mathcal{L}_{SPIB} = \frac{1}{N}\sum_{n=1}^{N}\int d \mathbf{z} \, p_{\theta}(\mathbf{z}|\mathbf{X}^{n})\Big[-\log q_{\theta}(\mathbf{y}^{n+\tau}|\mathbf{z}) + \beta \log\frac{p_{\theta}(\mathbf{z}|\mathbf{X}^{n})}{r_{\theta}(\mathbf{z})}\Big]
\end{align}

We begin by decomposing the term involving the prior into a cross entropy:

\begin{align}
    &\mathcal{L}_{D-SPIB} \notag \\
    &= \frac{1}{N}\sum_{n=1}^{N} \mathbb{E}_{p_{\theta}(\mathbf{z}|\mathbf{X}^{n})} \Big[-\log q_{\theta}(\mathbf{y}^{n+\tau}|\mathbf{z}) + \beta \log p_{\theta}(\mathbf{z}|\mathbf{X}^{n}) - \beta \log r_{\theta}(\mathbf{z})\Big] \label{eq:decomp} \\
    &= \frac{1}{N}\sum_{n=1}^{N} \Big[\mathbb{E}_{p_{\theta}(\mathbf{z}|\mathbf{X}^{n})} \Big[-\log q_{\theta}(\mathbf{y}^{n+\tau}|\mathbf{z}) + \beta \log p_{\theta}(\mathbf{z}|\mathbf{X}^{n})\Big] \notag \label{eq:CE} \\
    & + \beta \cdot \mathrm{H}(p_{\theta}(\mathbf{z}|\mathbf{X}^{n}) ||r_{\theta}(\mathbf{z})) \Big] \\
    &= \frac{1}{N}\sum_{n=1}^{N} \Big[\mathbb{E}_{p_{\theta}(\mathbf{z}_0|\mathbf{X}^{n})} \Big[-\log q_{\theta}(\mathbf{y}^{n+\tau}|\mathbf{z}_0) + \beta \log p_{\theta}(\mathbf{z}_0|\mathbf{X}^{n})\Big] \notag \\
    & + \beta \cdot \mathbb{E}_{t \sim \mathcal{U}[0,1]} \Big[ \frac{g(t)^2}{2} \mathbb{E}_{p_{\theta}(\mathbf{z}_t, \mathbf{z}_0 | \mathbf{X}^n)} \Big[ || \nabla_{\mathbf{z}_t} \log p(\mathbf{z}_t | \mathbf{z}_0) - \nabla_{\mathbf{z}_t} \log r_{\theta}(\mathbf{z}_t) ||^2_2 \Big] \Big] \notag \\
    & + const. \Big] \label{eq:thm}
\end{align}

Where in Eq. \ref{eq:decomp}, we have separated the posterior and prior into two terms, and in Eq. \ref{eq:CE} we have expressed the expectation of the log-prior over the posterior as a cross-entropy. In Eq. \ref{eq:thm}, we rewrite the cross-entropy as a score-matching loss term, following the same relabeling of $\mathbf{z} \rightarrow \mathbf{z}_0$ and assumption of a fixed noising kernel $p(\mathbf{z}_t | \mathbf{z}_0)$ derived from the pair of SDEs defining a noising process to a Gaussian prior explained in Sec. \ref{sec:architecture}. We note the existence of constant terms dependent on the noise schedule relevant for calculating the exact cross-entropy. Since the terms are not optimized we ignore them, giving Eq. \ref{eq:dspib}.

The expectation over the new joint distribution decomposes as follows: $p_{\theta}(\mathbf{z}_t, \mathbf{z}_0 | \mathbf{X}^n) = p(\mathbf{z}_t | \mathbf{z}_0)p_{\theta}(\mathbf{z}_0|\mathbf{X}^{n})$. Before going any further, we choose our $f(\mathbf{z},t)$ and $g(t)$ such that the diffusion process corresponds to the variance-preserving SDE (VP-SDE) commonly used in Denoising Diffusion Probabilistic Models (DDPMs):

\begin{align}
    f(\mathbf{z},t) &= - \frac{1}{2} \beta_{noise}(t)\mathbf{z} \\
    g(t) &= \sqrt{\beta_{noise}(t)} \label{eq:gnoise} \\
    \Rightarrow d \mathbf{z} &= - \frac{1}{2} \beta_{noise}(t)\mathbf{z} dt + \sqrt{\beta_{noise}(t)} d \mathbf{w}
\end{align}

where $\beta_{noise}(t)$ is an arithmetic progression from $\beta_{noise}(0)=1e-4$ to $\beta_{noise}(1)=0.2$; the values are chosen so that the final noised distribution closely approximates a Gaussian distribution. In practice, we discretize the SDE as a Markov chain, as done in DDPM. By the closure of Gaussian distributions and the reparameterization trick, we may explicitly draw noised samples at any arbitrary time in our forward process via:

\begin{align}
    \mathbf{z}_t = \sqrt{{\bar{\alpha}}_t} \mathbf{z}_0 +(1-\bar{\alpha}_t) \epsilon, \ \ \epsilon \sim \mathcal{N}(0,I)
\end{align}

where $\alpha_t = 1 - \beta_t$ and $\bar{\alpha}_t = \prod_{s=1}^t \alpha_s$. For the VP-SDE, the score is directly related to the noise via:

\begin{align}
    \nabla_{\mathbf{z}_t} \log p(\mathbf{z}_t|\mathbf{z}_0) = - \frac{\epsilon}{\sigma_t} = - \frac{\epsilon}{\sqrt{1 - \bar{\alpha}_t}} \label{eq:scoretonoise}
\end{align}

We arrive at our final loss function by expressing the score-matching objective as a noise-prediction one (they are related by the noise schedule), and approximating the expectation over the posterior distribution as a sum over data. We use the `unweighted' version of the denoising loss, in which $g(t)^2/\sigma_t^2 = 1$, i.e.,

\begin{align}
    &\mathcal{L}_{D-SPIB} \notag \\
    &= \frac{1}{N} \sum_{n=1}^N \Big[ \mathbb{E}_{p_{\theta}(\mathbf{z}_0|\mathbf{X}^{n})} \Big[ -\log q_{\theta}(\mathbf{y}^{n+\tau} | \mathbf{z}_0) + \beta \log p_{\theta}(\mathbf{z}_0 | \mathbf{X}^n) \Big] \notag \\
    &\quad + \beta \cdot \mathbb{E}_{t \sim U[0,1]} \Big[ \frac{1}{2} \frac{g(t)^2}{\sigma_t^2} \mathbb{E}_{\substack{p_{\theta}(\mathbf{z}_t, \mathbf{z}_0 | \mathbf{X}^n) \\ \epsilon \sim \mathcal{N}(0,I)}} \Big[ || \epsilon - \epsilon_{\theta}(\mathbf{z}_t, t)||^2_2 \Big] \Big] \Big] \\
    &= \frac{1}{N} \sum_{n=1}^N \Big[ \mathbb{E}_{p_{\theta}(\mathbf{z}_0|\mathbf{X}^{n})} \Big[ -\log q_{\theta}(\mathbf{y}^{n+\tau} | \mathbf{z}_0) + \beta \log p_{\theta}(\mathbf{z}_0 | \mathbf{X}^n) \Big] \notag \\
    &\quad + \beta \cdot \mathbb{E}_{t \sim U[0,1]} \Big[ \frac{1}{2} \mathbb{E}_{\substack{p_{\theta}(\mathbf{z}_t, \mathbf{z}_0 | \mathbf{X}^n) \\ \epsilon \sim \mathcal{N}(0,I)}} \Big[ || \epsilon - \epsilon_{\theta}(\mathbf{z}_t, t)||^2_2 \Big] \Big] \Big]
\end{align}

While this does not follow exactly from Eqs. \ref{eq:gnoise} and \ref{eq:scoretonoise}, we empirically observed that this choice resulted in more stable training.

\section{Training and sampling} \label{Algorithms}

Training of D-SPIB follows the standard SPIB training procedure (i.e., a self-consistent and iterative training approach) outlined in \cite{wang2021state}, with the exception of the new loss function. Training begins with multiple iterations of SPIB training, so that the latent space converges to a small number of relevant metastable states before the D-SPIB loss function is introduced. Otherwise convergence is hampered since the diffusion model is learning a `moving target'. The D-SPIB loss is entirely compatible with the vanilla SPIB training, during which short-lived states are merged or dropped when irrelevant, as only the decoder matters for this task.

For multi-temperature training, the variance of the Gaussian prior target is scaled by the temperature value in both training and sampling. Temperature information is also provided to the denoising network via learned embeddings. Samples are generated by simulating the VP-SDE (see Alg. \ref{alg:training}).

\begin{algorithm}
\caption{D-SPIB Sampling}\label{alg:training}
\begin{algorithmic}
\Require Trained $\epsilon_{\theta}(\mathbf{z}_t,t)$, samples $\mathbf{z}_1 \sim p_1(\mathbf{z_1})$, noise schedule $\beta_t$, diffusion timesteps $T_{diff}$, optional temperature $T$
\State $\alpha_t = 1 - \beta_t$
\State $\bar{\alpha}_t = \prod_{s=1}^t \alpha_s$
\State $t \gets T_{diff}$
\State $\mathbf{z}_T \gets \mathbf{z}_1$
\If{$T$ exists}
    \State $\sigma^2 = T^2$
\Else
    \State $\sigma^2 = 1$
\EndIf
\While{$t > 0$}
    \State $\mathbf{\gamma} \sim \mathcal{N}(0,\sigma^2I)$
    \State $\mathbf{z}_{t-1} \gets \frac{1}{\sqrt{\alpha_t}}(\mathbf{z}_t - \frac{1-\alpha_t}{\sqrt{1-\bar{\alpha}_t}}\epsilon_{\theta}(\mathbf{z}_t,t)) + \sqrt{\beta_t}\cdot \gamma$
\EndWhile
\State $\mathbf{z}_{0} \gets \frac{1}{\sqrt{\alpha_1}}(\mathbf{z}_1 - \frac{1-\alpha_1}{\sqrt{1-\bar{\alpha}_1}}\epsilon_{\theta}(\mathbf{z}_1,1))$
\State \textbf{return} $\mathbf{z}_0$
\end{algorithmic}
\end{algorithm}

\section{Experimental details} \label{Experimental Details}

In all experiments, the noise prediction network is a 7-layer fully connected sequential network of \texttt{torch.nn.Linear} layers, with \texttt{torch.nn.ReLU} activation functions. Diffusion time and simulation temperature are both embedded into the network by first taking Fourier feature projections and appending these features to the representation at each layer. Model hyperparameters for the two experiments in this work are listed in Table \ref{table:hyperparams}.

The two-dimensional three-hole potential is defined as follows:
\begin{small}
\begin{align}
    V(x, y) = 3e^{-x^{2}-(y-\frac{1}{3})^{2}} - 3e^{-x^{2}-(y-\frac{5}{3})^{2}} - 5e^{-(x-1)^{2}-y^{2}} - 5e^{-(x+1)^{2}-y^{2}} + 0.2x^{4} + 0.2(y-\frac{1}{3})^{4}
\end{align}
\end{small}
Molecular dynamics simulation of a single particle with mass $m=1$ is performed using the Langevin middle integrator. The system temperature is controlled at $1/k_{B}$, and reflective periodic boundary conditions were applied. The simulation is run for $5\times 10^{7}$ integration steps, with particle coordinates recorded every 50 steps. 

The Lennard-Jones 7 system consists of seven particles interacting via the Lennard-Jones potential in two dimensions. Six simulations were performed with a Langevin thermostat at temperatures from $0.2\epsilon/k_{B}$ to $0.7\epsilon/k_{B}$ in increments of $0.1\epsilon/k_{B}$. Each simulation ran for $10^{7}$ steps, with particle coordinates recorded every 100 steps. Additional simulation details for these two systems can be found in Ref \citep{qiu2025latent}.

\begin{table}
  \caption{Training Hyperparameters}
  \label{table:hyperparams}
  \centering
  \begin{tabular}{lcc}
    \toprule
    & \multicolumn{1}{c}{LJ7} & \multicolumn{1}{c}{Three well} \\
    \cmidrule(lr){2-2} \cmidrule(lr){3-3}
    Hyperparameter & Value & Value \\
    \midrule
    Lag Time & 1  & 20 \\
    Latent Dim. & 2  & 2 \\
    Encoder Type & Linear & Linear \\
    Batch Size & 512 & 512 \\
    Tolerance & 0.001 & 0.001 \\
    Patience & 5 & 5 \\
    Refinements & 10 & 10 \\
    Diffusion Patience & 150 & 50 \\
    Diffusion Refinements & 0 & 0 \\
    Random Seed & 42 & 42 \\
    Learning Rate & 0.001 & 0.001 \\
    Information Bottleneck $\beta$ & $1\times10^{-5}$ & $1\times10^{-5}$ \\
    Diffusion Steps & 100 & 100 \\
    \bottomrule
  \end{tabular}
\end{table}

\end{document}